\documentclass[runningheads]{llncs}
\usepackage{graphicx}
\usepackage{amsmath,amsfonts,amssymb}

\usepackage{comment}
\begin{document}
\title{Blind deblurring of hyperspectral document images\thanks{This project has received funding from the European Union's Horizon 2020 research and innovation programme under grant agreement No. 101026453. This work is published in the Lecture Notes in Computer Science book series (LNCS, volume 13373) as part of the  Image Analysis and Processing, ICIAP 2022 Workshops. DOI 10.1007/978-3-031-13321-3\_14}}
%
%
\author{Marina Ljubenovi\'c\inst{1}\orcidID{0000-0002-4404-3630} \and
Paolo Guzzonato\inst{1,2}\orcidID{0000-0002-0198-3008} \and
Giulia Franceschin\inst{1}\orcidID{0000-0003-1817-2962}\and
Arianna Traviglia\inst{1}\orcidID{0000-0002-4508-1540}}
\authorrunning{M. Ljubenovi\'c et al.}
%
\institute{Center for Cultural Heritage Technology, Istituto Italiano di Tecnologia, Venice, Italy \and
Ca'Foscari University of Venice, Venice, Italy\\
\email{\{marina.ljubenovic,paolo.guzzonato, giulia.franceschin,arianna.taviglia\}@iit.it}}
\maketitle              
\begin{abstract}
Most computer vision and machine learning-based approaches for historical document analysis are tailored to grayscale or RGB images and thus, mostly exploit their spatial information. Multispectral (MS) and hyperspectral (HS) images contain, next to the spatial information, a much richer spectral information than RGB images (usually spreading beyond the visible spectral range) that can facilitate more effective feature extraction, more accurate classification and recognition, and thus, improved analysis. 
Although utilization of a rich spectral information can improve historical document analysis tremendously, there are still some potential limitations of HS imagery such as camera induced noise and blur that require a carefully designed preprocessing step. Here, we propose novel blind HS image deblurring methods tailored to document images. We exploit a low-rank property of HS images (i.e., by projecting a HS image to a lower dimensional subspace) and utilize a text tailor image prior to performing a PSF estimation and deblurring of subspace components.
The preliminary results show that the proposed approach gives good results over all spectral bands, removing successfully image artefacts introduced by blur and noise and significantly increasing the number of bands that can be used in further analysis.

\keywords{Document images  \and Hyperspectral image processing \and Deblurring.}
\end{abstract}
%
%

\section{Introduction}
\label{sec:intro}

Automatic historical document analysis includes machine learning, computer vision, and pattern recognition based approaches for converting document images into a form that is easier for further manipulation (e.g., storage, systematic evaluation, information retrieval, and forensic analysis). Some of the common tasks in historical document analysis include layout analysis \cite{1998_Cattoni_LayoutAnalysis}, optical character recognition \cite{1990_Govindan_OCR,2020_Kumar_AncientOCR}, automatic transcription and translation \cite{2009_Fischer_Transcription}, and information retrieval \cite{2013_Hedjem_DocRestoration}.
Most computer vision and machine learning-based approaches are tailored to grayscale or RGB images and thus, mostly exploit their spatial information (i.e., locations of the pixels in the spatial domain). Multispectral (MS) and hyperspectral (HS) images contain, next to the spatial information, much richer spectral information than RGB images (usually going beyond the visible spectral range). This information can further improve image analysis by facilitating more effective feature extraction and more accurate classification and recognition. 

\noindent HS images are usually represented in a form of a 3-dimensional (3D) data cube, where 2D spectral bands corresponding to different wavelengths are stacked together along the spectral axis. These images are most commonly used in remote sensing applications \cite{2007_Govender_HSIReview}. In recent years, closed range HS images gain popularity in other computer vision applications in agriculture \cite{2015_Mahesh_HSAgriculture}, chemistry \cite{2012_Kamruzzaman_HSChemistry}, and cultural heritage \cite{2020_Picollo_HSI_CH}. HS imagery has been applied to tackle different problems in historical document analysis such as material identification and inks characterization \cite{2014_Shiradkar_MaterialClassification}, and removal of ageing-related artefacts (e.g., foxing and ink bleeding) \cite{2011_Kim_OldDoc}.
It is shown in the literature that some materials used for the preparation of ancient documents (i.e., inks, supports, and bindings) have a unique fingerprint in the hyperspectral domain \cite{2015_George_InkClassification}. This fact can be utilized for material identification and digital source separation (e.g., separation of foxing artefacts or stains from inks) and ameliorate further image processing and analysis. 
Although utilization of rich spectral information can improve historical document analysis tremendously, there are still some potential limitations of HS imagery, such as camera induced noise and blur, that require a carefully designed preprocessing step. 

\noindent Over years, many HS image restoration approaches were proposed, mostly tailored to MS and HS remote sensing images \cite{2015_Loncan_PasharpeningReview,2018_Zhuang_FastHyDe}, where the low spatial resolution represents a major limitation for further data processing and analysis (e.g., satellite images usually have a spatial resolution of more than $10 \times 10$ m per pixel). For document image processing, where HS images are taken in the close range, image degradation induced by a camera, light conditions, and acquisition settings can corrupt all or only some spectral bands, making them unusable. These image degradation are often unavoidable and thus, prior denoising and deblurring represent a necessary preprocessing step for HS document image analysis. 

\noindent To remove noise from HS images, the majority of state-of-the-art methods exploit characteristics of the HS data in the spectral domain such as a low-rank representation \cite{2018_Zhuang_FastHyDe}. The low-rank assumption is widely utilized for other tasks such as spectral unmixing \cite{2013_Zhao_Unmix_Deb}, classification \cite{2012_Fang_Coupled}, and super-resolution \cite{2017_Lanares_SuperResolution}. These tasks are commonly coupled with HS image deblurring where the point-spread function (PSF) of the imaging system is considered to be known and constant over bands. Additionally, remote sensing methods are designed for natural images, thus utilizing priors adjusted to these types of images. It is shown in the literature that images containing text follow different statistics compared to natural images (e.g., often contain only two colours and sharp transitions) and thus, require carefully designed or learned image priors \cite{2014_Pan_TextDeb,2019_Ljubenovic_IJDAR}. To learn an image prior, methods based on Gaussian mixture models \cite{2017_Ljubenovic_ICIP} and specially designed dictionaries \cite{2017_Ljubenovic_ICDAR} have been proposed. These methods, tailored to RGB document images, do not exploit the high correlation between spectral bands: they can be applied to each band separately, leading to a higher sensitivity to the presence of noise and a time-consuming restoration process.
Methods based on neural networks are also exploited in the past for text deblurring \cite{2015_Hradis_CNNText}. These methods, although providing good results on noiseless RGB images, are sensitive to even a small amount of noise (commonly present in bands of HS images corresponding to blue and near-infrared spectral regions).

\noindent Other remote sensing approaches, focused solely on deblurring, exploit different dimensionality reduction methods (e.g., \textit{principal component analysis} \cite{2013_Liao_HSIDeblurring} or \textit{singular value decomposition} (SVD) \cite{2021_Ljubenovic_SPIE}) to tackle blur present in HS images. Dimensionality reduction (i.e., projection of higher dimensional data to a lower-dimensional subspace) is a powerful tool as it allows for better differentiation between different spectral features. Additionally, by projecting data to a lower-dimensional subspace, useful information is usually preserved and noise reduced.  Dimensionality reduction approaches are successfully used for the identification of endmembers (i.e., pure materials) in HS remote sensing data \cite{2005_Nascimento_VCA}.
In document images, the number of endmembers is often limited (e.g., a document usually contains three materials: ink, support, and binding). Also, document HS images are particularly sensitive to poor light settings during acquisition which can introduce additional artefacts (e.g., shades).

\noindent Here, we propose a novel blind HS image deblurring method tailored to document images. We exploit a low-rank property of HS images and utilize a text tailor image prior to performing a PSF estimation and deblurring of subspace components. Specially designed samples containing lab-created and commercial inks on a printing paper are used to evaluate the performance of the proposed approach.  

\section{Proposed method}
\label{sec:method}

Assuming additive noise, hyperspectral image deblurring is modelled as 
\begin{equation}
\label{equ:model_blur}
\textbf{Y} = \textbf{HX} + \textbf{N},
\end{equation}
where $\textbf{Y} \in \mathbb{R}^{b \times n}$ and $\textbf{X} \in \mathbb{R}^{b \times n}$ represent an observed and an underlying hyperspectral images, respectively, with the rows containing $b$ spectral bands. Every band is a vectorized image containing $n$ pixels. $\textbf{H}$ and $\textbf{N} \in \mathbb{R}^{b \times n}$ represent a blurring matrix and a Gaussian independent and identically distributed (i.i.d.) noise, respectively. Blurring matrix has the block-circulant-circulant-block (BCCB) form, where each block depict the cyclic convolution associated with a camera point spread function (PSF). 

\noindent The spectral vectors $\textbf{x}_i$, for $i = 1,...,n$, live in a lower-dimensional subspace $\mathcal{S}_k$, where the number of dimensions $k \ll b$ \cite{2018_Zhuang_FastHyDe} and thus, the underlying image $\textbf{X}$ can be represented as

\begin{equation}
\textbf{X} = \textbf{E} \textbf{Z},
\label{equ:x_ez}
\end{equation}
where $\textbf{E} = [\textbf{e}_1,...,\textbf{e}_k] \in \mathbb{R}^{b \times k}$ stands for the subspace bases and $\textbf{Z} \in \mathbb{R}^{k \times n}$ holds the representation coefficients of $\textbf{X}$ in $\mathcal{S}_k$.

\noindent In the proposed method, we start by projecting the observed HS image to a lower dimension subspace by applying a state-of-the-art dimensionality reduction method (HySime) \cite{2007_Nascimento_HySime}. Then, we perform blind deblurring on each subspace component separately by using a text tailored image prior.

\noindent To estimate each sharp subspace component and blurring operator corresponding to that component, we solve the following optimization problem 

\begin{equation}
\label{equ:prob_blur_eigen}
\hat{\textbf{z}}_i, \hat{\textbf{h}}_i \in
\underset{\textbf{z}_i, \textbf{h}_i}{\mathrm{argmin}} 
\frac{1}{2} ||\textbf{y}_i - \textbf{H}_i \textbf{E} \textbf{z}_i||_2^2 +
\lambda \Phi(\textbf{z}_i) +
\gamma ||\textbf{h}_i||_2^2,
\end{equation}
where $\Phi$ represents a prior knowledge imposed to the underlying sharp image and $\lambda$ and $\gamma$ are regularization parameters that control a trade off between the data fidelity term and priors, respectively.

\noindent Following the formulation from \cite{2014_Pan_TextDeb}, the image prior imposed on subspace components is formulated as a sum of two so-called $l_0$-norms on image intensities and gradients with the following form
\begin{equation}
\label{equ:image_prior}
\Phi(\textbf{z}_i) = \alpha||\textbf{z}_i||_0 + ||\nabla\textbf{z}_i||_0,
\end{equation}
with $\alpha$ controlling the contribution of each component. 

\noindent To solve the optimization problem from (\ref{equ:prob_blur_eigen}), we first fix the image $\textbf{z}_i$ and estimate the blurring operator (by solving the least square minimisation). When blurring operator estimate is obtained, we apply a half-quadratic splitting optimization approach to obtain the sharp image (for more details about the optimization procedure, take a look at \cite{2011_Xu_ImageSmoothing} and references therein). 

\section{Experiments}
\label{sec:exp}

To test the proposed HS image deblurring method, we created several samples by using lab-created and commercial inks and a printing paper. 

\begin{figure} [ht]
   \begin{center}
   \includegraphics[width=0.95\textwidth]{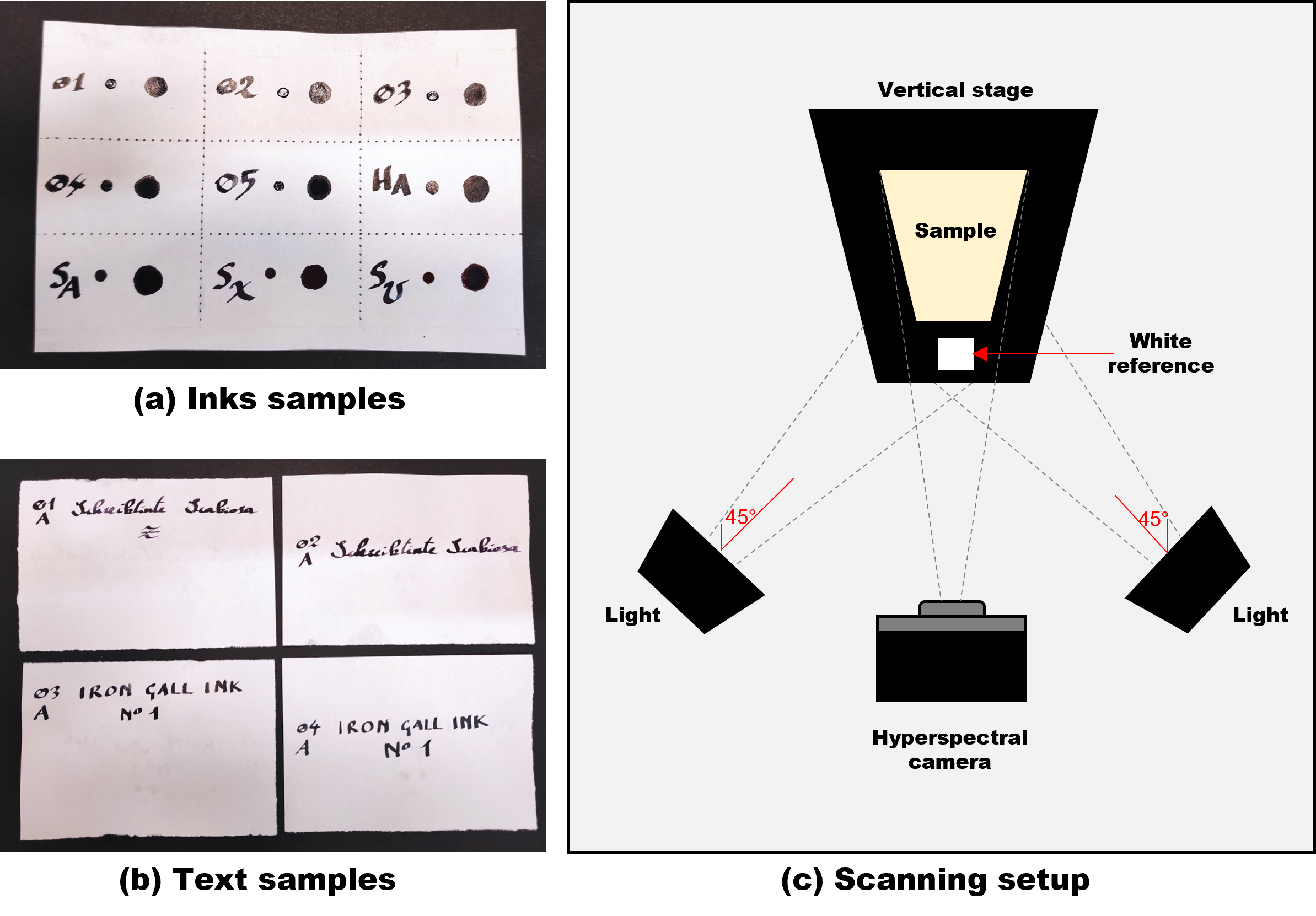}
   \end{center}
   \caption{(a) Samples of different commercial and lab-created inks; (b) samples of text written with these inks; and (c) an illustration of the acquisition setup with a HS camera and two led lamps.}
   \label{fig:samples}
\end{figure}

\noindent \textbf{Lab-created inks.} Five iron gall inks were prepared with simplified recipes based on the Madrid ink of Díaz Hidalgo \textit{et al.} \cite{2018_Diaz_IronGall}. Inks have different gall:FeSO4 weight ratio, namely 1:1 for the first ink (Fig. \ref{fig:samples}(a), ink 01), 3:2 for the second ink (02), and 2:1 for the others (03, 04, and 05). 
For inks 01 to 03 iron sulfate was added to a solution obtained macerating galls powder in water for a week.
For the ink 04 iron sulfate was grounded with the galls and macerated in water for a week.
Ink 05 is akin to ink 03, but with the process shortened to one day. 
All the inks were filtered and contain the same amount of Arabic gum and water. Galls (Q. infectoria), iron(II) sulfate heptahydrate, and Arabic gum powder (A. senegal) were purchased from Kremer. 

\noindent \textbf{Commercial inks.} Commercial iron based inks ($H_A$, $S_A$, $S_X$) and copper based ink ($S_V$) were used as references. The commercial inks are: 1) Jacques Herbin Encre Authentique ($H_A$), 2) Roher \& Klinger  Schreibtinte Eisen-Gall-Tinte Scabiosa ($S_A$), 3) Roher \& Klinger Schreibtinte Eisen-Gall-Tinte Salix ($S_X$), 4) Roher \& Klinger Verdigris ($S_V$).

\noindent Samples of ink on (80 $g / m^2$) printing paper were written using steel and glass nibs. Inks spots were made with 10 $\mu l$ spread and unspread drops.

\noindent The HS images are obtained by a hyperspectral camera based on time-domain Fourier Transform detection (NIREOS HERA Iperspettrale VIS-NIR). The camera has a spectral range from 400 to 1000 nm and the spatial resolutions $1280 \times 1024$ pixels. The camera was placed around 1.5 m from the sample that was mounted on a vertical stage. Two led lamps with a blue filter (5600k) were set on each side of the camera, illuminating a sample at an angle of $45^{\circ}$ (Fig. \ref{fig:samples}(c)).

\section{Results and Discussion}
\label{sec:results}

We compared the proposed method with two state-of-the-art approaches for HS image deblurring: 1) The method that exploits the PCA dimensionality reduction approach and performs \textit{total variation} (TV)-based deblurring only on the first few subspace components (PCA+TV) \cite{2013_Liao_HSIDeblurring}; 2) The method based on the HySime approach followed by deblurring of subspace components tailored to remote sensing HS images (RS HS deblurring) \cite{2021_Ljubenovic_SPIE}. For a fair comparison, we selected only methods tailored to HS images and not the ones developed for RGB or grayscale images as the latest may be sensitive to even a small amount of noise, often present in some HS bands.

\begin{figure} [ht]
   \begin{center}
   \includegraphics[width=0.95\textwidth]{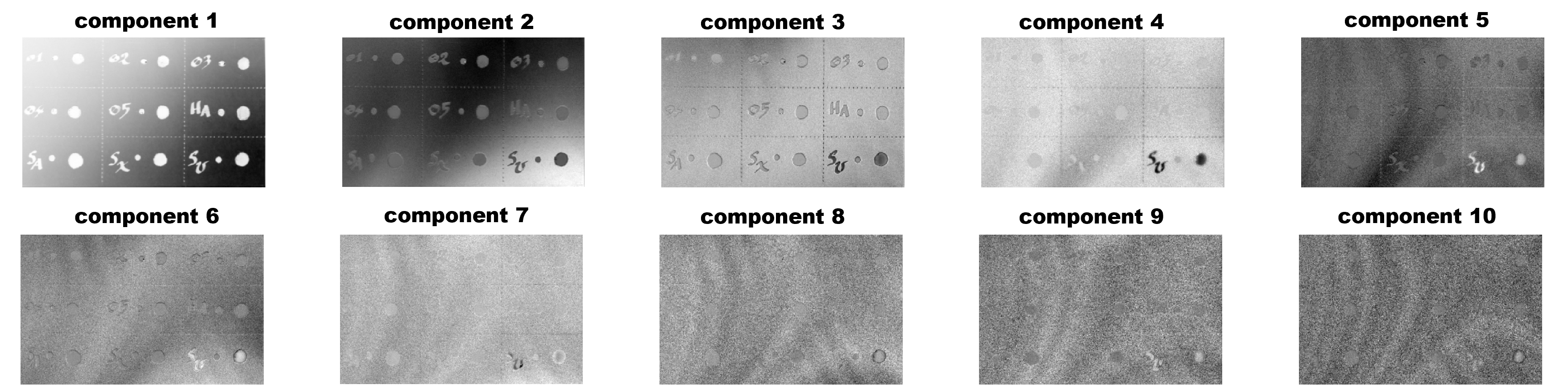}
   \end{center}
   \caption{Example of subspace components of a sample containing nine different inks.}
   \label{fig:subspace_components}
\end{figure}

\noindent The first step of the proposed method is dimensionality reduction: a HySime approach is applied to the observed HS image \cite{2007_Nascimento_HySime} to obtain subspace components. Fig. \ref{fig:subspace_components} shows the first 10 subspace components of a sample containing nine different inks (five lab-created and four commercial). The useful information is visible in the first five subspace components, while the next five components contain mostly noise. The copper-based commercial ink ($S_V$) shows different behaviour in the subspace domain when compared with iron gall inks (lab-created and commercial). 

\begin{figure} [ht]
   \begin{center}
   \includegraphics[width=0.95\textwidth]{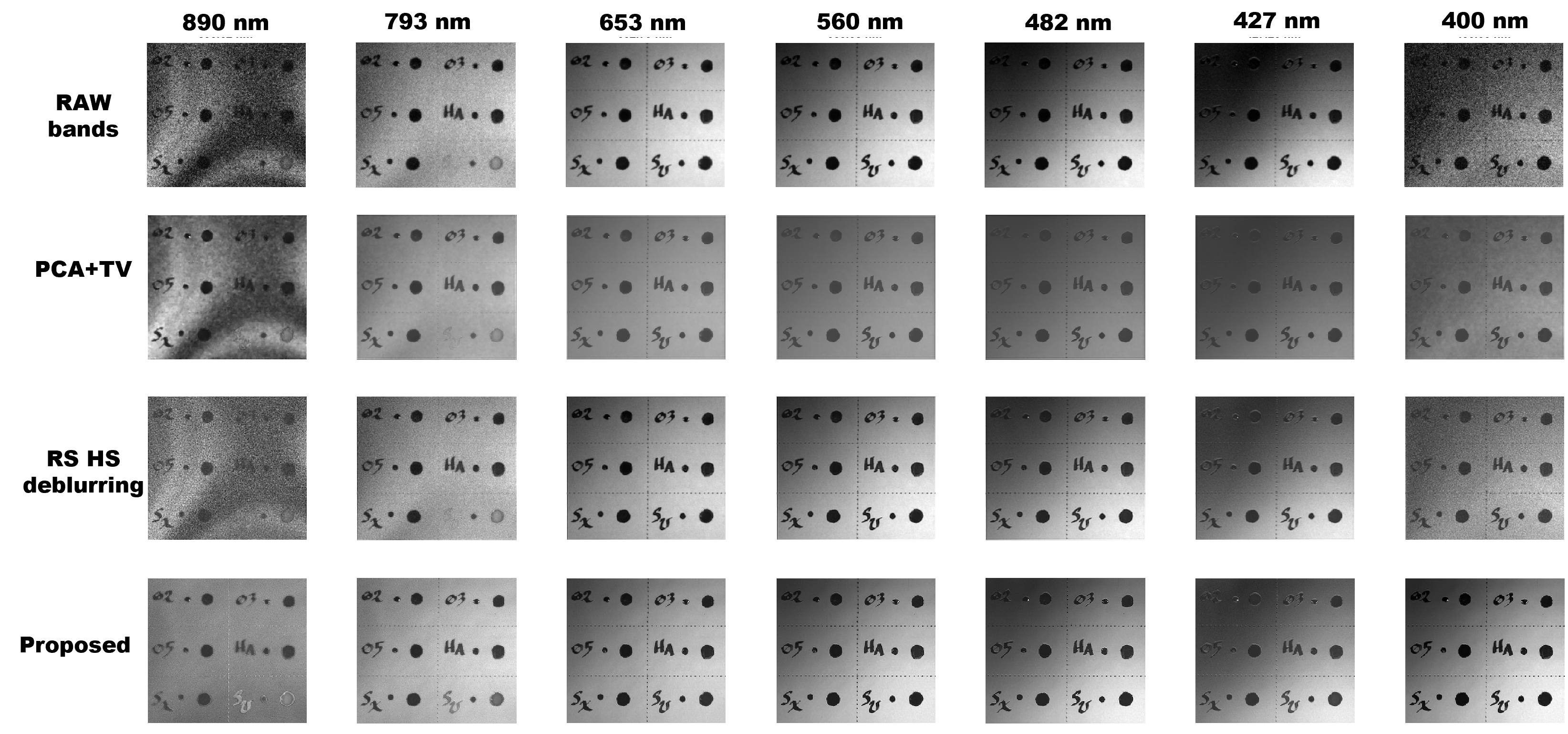}
   \end{center}
   \caption{Comparison of the results obtained by three approaches: the PCA and TV-based method (PCA+TV) \cite{2013_Liao_HSIDeblurring}, the deblurring method tailored to remote sensing images (RS HS deblurring) \cite{2021_Ljubenovic_SPIE}, and the proposed method.}
   \label{fig:bands}
\end{figure}

\noindent The results obtained on the separate bands of HS document images are presented in Fig. \ref{fig:bands}. All three tested deblurring methods perform reasonably well on bands corresponding to medium frequencies (e.g., from 653 to 427 nm). PCA+TV introduces new artefacts on noisy bands (e.g., the band corresponding to 890 nm) and oversmooth small details in the image. Similarly, the method tailored to remote sensing images (RS HS deblurring) fails to fully remove noise from bands corresponding to the near-infrared (NIR) region ($\geq$ 793 nm). The proposed method performs optimally over the full spectral range and provides the sharpest resulting image. 

\begin{figure} [ht]
   \begin{center}
   \includegraphics[width=0.95\textwidth]{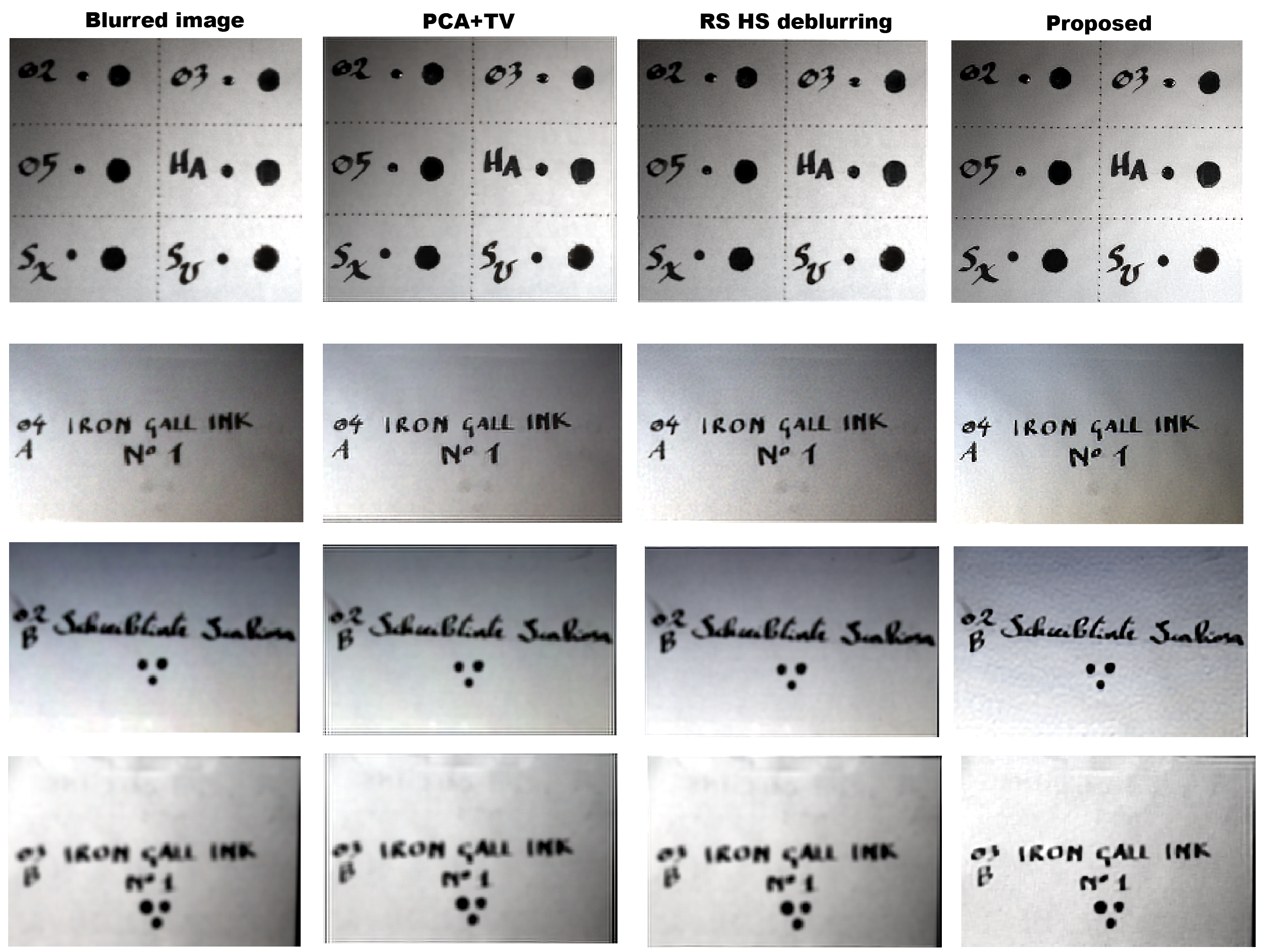}
   \end{center}
   \caption{RGB composition obtained on three different samples (blurred images presented in the first column); Methods: the PCA and TV-based method (PCA+TV) \cite{2013_Liao_HSIDeblurring}, the deblurring method tailored to remote sensing images (RS HS deblurring) \cite{2021_Ljubenovic_SPIE}, and the proposed method.}
   \label{fig:rgb}
\end{figure}

\noindent Finally, we show the results in the form of an RGB composition (Fig. \ref{fig:rgb}). As previously, we compared the results obtained by three HS image deblurring approaches. The three approaches give comparable results when tested on the images corrupted by a moderate blur (first and second rows), except the visible boundary artefacts visible when applying the PCA+TV approach. When an HS image is blurrier (third and fourth rows), the proposed method outperforms other tested approaches.  

\section{Conclusion}
\label{sec:conclusion}

In this paper, we present a novel blind deblurring method tailored to hyperspectral images that contain text. The proposed method is based on the well-known low-rank characteristic of HS data and exploits a text-specific prior knowledge imposed on an underlying sharp HS image. The image prior is motivated by observing specific properties of text images.
The behaviour of different lab-created and commercial inks in the visible and near-infrared spectral ranges was investigated to improve the restoration process. 
The preliminary results show that the proposed approach gives good results over all spectral bands, removing successfully image artefacts introduced by blur and noise and significantly increasing the number of bands that can be used in further analysis.

\noindent The proposed approach represents a (crucial) preprocessing step for historical document analysis where the important information may be hidden in the spectral ranges beyond visible light. Additionally, the proposed approach gives a new insight into the characterisation and identification of ancient inks.
The future work will be focused on a more detailed examination of the behaviour of ancient materials in NIR and short-wave infrared wavelength ranges, together with the utilization of these ranges in the process of HS image restoration.
\bibliographystyle{splncs04}
\bibliography{hsi}

\end{document}